%% file: main_v2_arxive.tex
\documentclass[runningheads]{llncs}

\usepackage[FINAL,year=2026,ID=*****]{eccv}

\usepackage{eccvabbrv}

\usepackage{graphicx}
\usepackage{booktabs}
\usepackage[acronym]{glossaries}
\makeglossaries
\glsdisablehyper

\usepackage[accsupp]{axessibility}  

\usepackage{hyperref}

\usepackage{orcidlink}

\begin{document}
\include{acronyms}
\title{Team RAS in 11th ABAW Competition:\\
Multimodal Ambivalence Recognition Approach
} 

\titlerunning{Team RAS: Multimodal Ambivalence Recognition Approach}

\author{
Elena Ryumina\inst{1}\orcidlink{0000-0002-4135-6949} \and
Maxim Markitantov\inst{1}\orcidlink{0000-0001-7987-1025} \and
Alexandr Axyonov\inst{1}\orcidlink{0000-0002-7479-2851} \and \\
Fedor Shchetinin\inst{2}\orcidlink{0009-0006-0648-3105} \and
Timur Abdulkadirov\inst{1}\orcidlink{0009-0007-9439-1813} \and
Dmitry Ryumin\inst{1}\orcidlink{0000-0002-7935-0569} \and \\
Alexey Karpov\inst{1,3}\orcidlink{0000-0003-3424-652X}
}

\authorrunning{E.~Ryumina et al.}

\institute{
St. Petersburg Federal Research Center of the Russian Academy of Sciences \\ 
(SPC RAS), St. Petersburg, Russia\\
\email{\{ryumina.e,markitantov.m,axyonov.a,ryumin.d,karpov\}@iias.spb.su}
\email{timur.abdulkadirov4@gmail.com}
\and
HSE University, St. Petersburg, Russia\\
\email{fshchetinin@hse.ru}\\
\and
ITMO University, St. Petersburg, Russia
}

\maketitle

\begin{abstract}
  Automatic recognition of ambivalence and hesitancy is challenging because these states may be expressed through inconsistent linguistic, acoustic, facial, and contextual patterns, while top-performing systems often rely on computationally expensive ensembles. We present a single text-centered multimodal approach for video-level ambivalence and hesitancy recognition for the 11th \acrfull{ABAW} Challenge. The proposed approach combines linguistic, acoustic, facial, and scene features using
  text-centered
multimodal fusion model. Text Residual Fusion treats text as the anchor modality and applies gated residual adjustments based on the other modalities. 
  Experiments on the \acrfull{BAH} corpus confirm that text is the strongest unimodal modality. The Text Residual Fusion model achieves an average \acrfull{MF1} of 75.14\% across the Development and Public Test subsets. On the Private Test subset, it reaches an \acrshort{MF1} of 78.24\%, outperforming the text model by 4.03\%. These results demonstrate that complementary multimodal information can improve recognition performance without requiring a large model ensemble.
\end{abstract}

\section{Introduction}
\label{sec:intro}

Ambivalence and hesitancy are complex affective states that may be expressed through inconsistent linguistic, acoustic, facial, and contextual information. Their automatic recognition is therefore relevant to affective computing, which studies human affect from face~\cite{sajjad2023comprehensive}, audio~\cite{george2024review}, text~\cite{deng2021survey}, and bodily modalities~\cite{leong2023facial}, with applications in human--computer interaction, healthcare, education, and assistive systems~\cite{poria2017review}. The \gls{AH} Video Recognition task of the 11th \gls{ABAW} Challenge addresses this problem at the video-level using the \gls{BAH} corpus~\cite{gonzalezbah}. In digital behavior-change scenarios, such states may reflect uncertainty, resistance, unstable motivation, or potential disengagement~\cite{bijkerk2024engagement}.

Previous studies have shown that transcripts provide the strongest unimodal signal for this task, whereas facial and acoustic information can further improve performance when fused effectively~\cite{gonzalezbah,Savchenko_2025_CVPR,Hallmen_2025_CVPR}. This finding motivates an asymmetric fusion strategy in which text serves as the primary semantic representation, while audio, face, and scene modalities provide complementary information. This design is consistent with recent studies on modality-aware and uncertainty-aware multimodal fusion~\cite{fang2025emoe}, as well as with broader advances in multimodal affective computing~\cite{kollias2025advancements,kollias2025emotions}.

The 10th \gls{ABAW} Challenge also highlighted a practical limitation of current top-performing systems~\cite{pereira2026brother,bekhouche2026conflict,ryumina2026team}: the three highest-ranked solutions used ensembles or multiple prediction branches, and the winning method combined approximately twenty models~\cite{pereira2026brother}. Although such systems may achieve high benchmark performance, they require substantially greater computational resources, inference time, memory, and implementation effort. In this work, we propose a single text-centered multimodal approach that combines text, audio, face, and scene modalities without relying on a large ensemble. The proposed approach aims to preserve complementary multimodal information while offering a more compact and practically deployable solution for video-level \gls{AH} recognition.

\section{Related Work}

\subsection{Ambivalence and Hesitancy Recognition}

The \gls{BAH} benchmark introduced in~\cite{gonzalezbah} established the main experimental setting for video-level \gls{AH} recognition. The authors evaluated unimodal and multimodal systems using face, audio, and text modalities. The examined fusion methods included feature concatenation, co-attention, Transformer-based fusion, and cross-attention. The results showed that text was the strongest individual modality. The official baseline also included a zero-shot \gls{M-LLM} based on Video-LLaVA~\cite{lin2024video}.

Submissions to the challenge confirmed this finding. Hallmen et al.~\cite{Hallmen_2025_CVPR} combined visual features extracted using a \gls{ViT}~\cite{caron2021emerging}, acoustic features obtained using Wav2Vec2~\cite{baevski2020wav2vec2}, and textual features produced by \gls{BERT}~\cite{bert2019}. They used \glspl{LSTM}~\cite{hochreiter1997lstm} and an \gls{MLP}~\cite{rumelhart1986learning} for temporal modeling and multimodal fusion. Savchenko et al.~\cite{Savchenko_2025_CVPR} extracted facial, acoustic, and textual features with EmotiEffLib~\cite{savchenko2024emotieffnet}, Wav2Vec2~\cite{baevski2020wav2vec2}, and RoBERTa~\cite{liu2019roberta}, respectively, and evaluated both early and late fusion strategies. Both studies identified text as the strongest unimodal source, while improvements from multimodal fusion depended on the effective preservation of complementary information~\cite{fang2025emoe}.

\subsection{Top Systems of 10th ABAW Challenge}

The three highest-ranked solutions in the \gls{AH} task of the 10th \gls{ABAW} Challenge relied on model ensembles. The third-ranked system~\cite{ryumina2026team} combined scene, facial, acoustic, and textual features. Its final results on the Private Test subset was obtained using an ensemble of five prototype-augmented multimodal fusion models. The second-ranked ConflictAwareAH system~\cite{bekhouche2026conflict} combined video, audio, and text modalities with pair-wise cross-modal conflict features and text-guided late fusion. Its final submission also used an ensemble of five models. The winning BROTHER system~\cite{pereira2026brother} employed a considerably larger ensemble of approximately twenty heterogeneous models and combined their predictions using a voting strategy.

Although these results demonstrate the effectiveness of model ensembles, the use of five to twenty models substantially increases computational cost, inference latency, memory requirements, and implementation complexity. These factors complicate the practical deployment and reproduction of such systems and motivate our focus on a single text-centered multimodal fusion model.

\section{Proposed Approach}

\begin{figure}[ht]
    \centering
    \includegraphics[width=\linewidth]{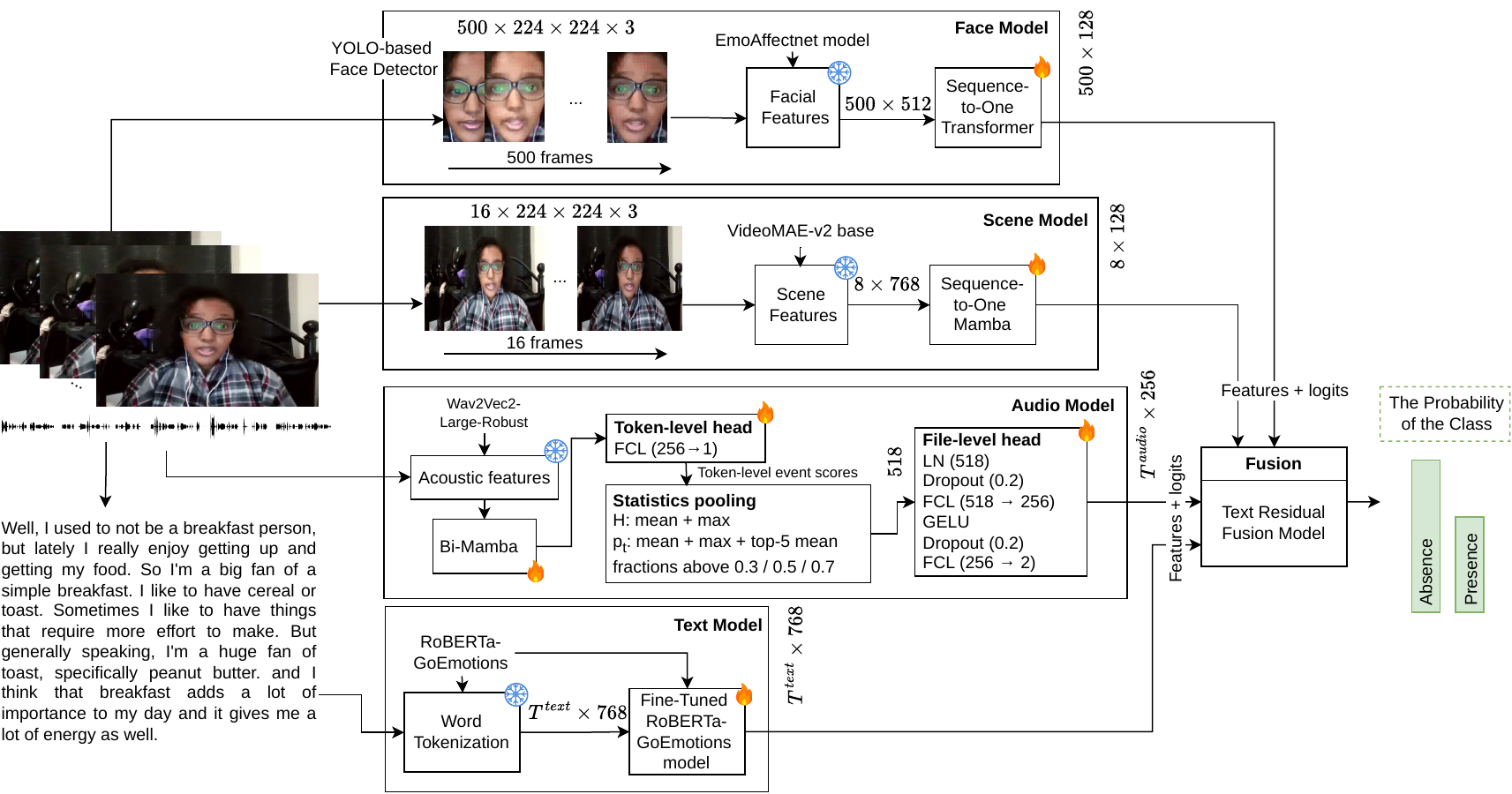}
    \caption{Pipeline of text-centered multimodal approach.}
    \label{fig:pipeline_approach}
\end{figure}

The pipeline of the proposed approach is shown in Figure \ref{fig:pipeline_approach}. The details are described below.

\subsection{Text Model}
Textual features are extracted using a RoBERTa-base model pre-trained on the GoEmotions dataset\footnote{\url{https://huggingface.co/SamLowe/roberta-base-go_emotions}}. During training, the embedding layer and the first four Transformer encoder layers are frozen.

Let an input text sequence of length $T^{text}$ be represented by a matrix of token embeddings:
\begin{equation}
X^{text} = [x^{text}_1, x^{text}_2, \ldots, x^{text}_{T^{text}}] \in \mathbb{R}^{T^{text} \times 768}.
\end{equation}

Each token $x^{text}_t$ is mapped to a contextual representation by the encoder layers, producing the hidden-state matrix:
\begin{equation}
H^{text} = \text{Encoder}(X^{text}) = [h^{text}_1, h^{text}_2, \ldots, h^{text}_{T^{text}}] \in \mathbb{R}^{T^{text} \times 768}
\end{equation}

The pooled representation corresponding to the classification token is used for the final prediction. This sentence-level representation is passed to an \gls{MLP} classification head consisting of a dropout layer, a \gls{FCL}, a $\tanh$ activation function, a second dropout layer, and a final \gls{FCL} with two output units.

\subsection{Audio Model}
Before training the audio model, each video is converted into a waveform sampled at 16~kHz and processed as a full audio, without voice-activity filtering. The acoustic features are extracted from the tenth Transformer layer of a frozen Wav2Vec2 model~\cite{baevski2020wav2vec2}\footnote{\url{https://huggingface.co/facebook/wav2vec2-large-robust}}. The resulting audio sequence is represented as
\begin{equation}
    X^\text{audio}
    =
    [x^\text{audio}_1, x^\text{audio}_2, \ldots,
    x^\text{audio}_{T^\text{audio}}]
    \in
    \mathbb{R}^{T^\text{audio} \times 1024},
\end{equation}
where $x^\text{audio}_t$ denotes the acoustic embedding corresponding to the $t$-th temporal interval. The extracted features are first projected into a compact hidden space and then processed by a bi-directional Mamba~\cite{gu2023mamba} block, followed by dropout and a residual connection. The resulting temporal acoustic features are denoted by $H^\text{audio}\in\mathbb{R}^{T^\text{audio}\times256}$.

Because \gls{AH} behavior may occur only during short intervals of a recording, frame-level annotations are used as auxiliary temporal supervision to help the model localize such events. These annotations are converted into token-level soft targets. For the set $\mathcal{F}_t$ of video frames temporally aligned with the $t$-th audio token, the target is defined as
\begin{equation}
    \widetilde{y}_t
    =
    \frac{1}{|\mathcal{F}_t|}
    \sum_{j\in\mathcal{F}_t} y_j.
\end{equation}

A token-level head consisting of a single \gls{FCL} produces a token-level logit $z_t$ and the corresponding probability $p_t=\sigma(z_t)$. For file-level classification, masked mean and max pooling over $H^\text{audio}$ are concatenated with the mean, maximum, top-$K$ mean, and threshold statistics of the token-level probabilities, producing a $518$-dimensional descriptor. This descriptor is processed by an \gls{MLP} consisting of two \glspl{FCL}. \gls{LN} and dropout are applied before the first layer, while a GELU activation function and dropout are applied between the two \glspl{FCL}. The final \gls{FCL} contains two output units. The audio model is trained using a combination of file-level and token-level supervision:
\begin{equation}
\mathcal{L}
=
\mathcal{L}_{\mathrm{file-level}}
+
0.5 \mathcal{L}_{\mathrm{token-level}}
+
0.02 \mathcal{L}_{\mathrm{smooth}}
+
0.1 \mathcal{L}_{\mathrm{event}}.
\end{equation}
where $\mathcal{L}_{\mathrm{file-level}}$ denotes the main cross-entropy loss for the file-level \gls{AH} label. The auxiliary term $\mathcal{L}_{\mathrm{token-level}}$ is a token-level binary cross-entropy loss whose soft targets are obtained from the 24 fps frame-level annotations. The smoothness loss penalizes isolated peaks in the token-level probabilities, while the event loss applies binary cross-entropy to the maximum token logit, encouraging the model to detect short \gls{AH} intervals within long recordings. The auxiliary-loss weights were selected based on performance of the Development subset. 

\subsection{Face Model}
Face regions are detected using a YOLO-based face detector\footnote{\url{https://github.com/lindevs/yolov8-face}}. The detected regions are resized to $224 \times 224$ pixels and processed by the frame-level facial emotion encoder Emo-AffectNet~\cite{RYUMINA2022435}, which produces a $512$-dimensional feature vector for each face. A video containing $T^{\text{face}}$ retained face regions is represented as
\begin{equation}
    X^\text{face} = [x^\text{face}_1, x^\text{face}_2, \ldots, x^\text{face}_{T^\text{face}}] \in \mathbb{R}^{T^\text{face} \times 512},
\end{equation}
where $x^\text{face}_t$ denotes the frame-level embedding. A maximum of $T^\text{face}=500$ frames is used. Longer sequences are uniformly subsampled, while shorter sequences are remained unchanged.

The temporal dynamics of facial \gls{AH} expressions are modeled with a Transformer-based encoder~\cite{vaswani2017attention}. The Emo-AffectNet embeddings are projected into a $128$-dimensional hidden space, combined with positional encodings, and processed by a single Transformer layer with eight attention heads. The temporally pooled representation is then passed to an \gls{MLP} classification head consisting of a \gls{FCL}, \gls{LN}, a GELU activation function, dropout, and a final \gls{FCL} with two units. The output of the penultimate \gls{FCL} is used as the facial classification features, and the output of the final layer represents the facial logits.

To regularize the model, we apply flow matching~\cite{jha2026discriminativeflowmatchinglocal} in both the temporal space and the logit spaces. Let $Z$ denote the temporal features produced by the Transformer-based encoder from the input sequence $X^\text{face}$. For feature-space flow matching, a noise tensor $Z_0$ is interpolated with $Z$ using $t \sim \mathcal{U}(0,1)$:
\begin{align}
    Z_t = (1-t) Z_0 + t Z, \mathcal{L}_{\mathrm{FM}}^{\mathrm{feat}} &=
    \left\| f_\theta(Z_t, Z, t) - (Z - Z_0) \right\|_2^2 .
\end{align}

For logit-space flow matching, the initial logits $\hat{y}_0$ are interpolated with the one-hot target vector $q$:
\begin{align}
    y_t = (1-t)\hat{y}_0 + t q, \mathcal{L}_{\mathrm{FM}}^{\mathrm{logit}} &=
    \left\| g_\psi(y_t, z, t) - (q-\hat{y}_0) \right\|_2^2 .
\end{align}

The final training loss is defined as
\begin{equation}
    \mathcal{L} =
    \mathcal{L}_{\mathrm{CE}} +
    \lambda_{\mathrm{FM}}
    \left(
    \mathcal{L}_{\mathrm{FM}}^{\mathrm{feat}} +
    \mathcal{L}_{\mathrm{FM}}^{\mathrm{logit}}
    \right),
\end{equation}
where $\mathcal{L}_{\mathrm{CE}}$ is the class-weighted cross-entropy loss for two-class classification. Flow matching is performed using four integration steps, with $\lambda_{\mathrm{FM}}=0.1$.

\subsection{Scene Model}
Scene information is extracted from the full video frames rather than from cropped face regions. We use a VideoMAE-v2-base visual encoder~\cite{wang2023videomae} as the scene feature extractor. Sixteen frames are uniformly sampled from the entire video to preserve information about the background, body posture, and interaction context. For each video, the encoder produces a sequence of embeddings:

\begin{equation}
    X^\text{scene} =
    [x^\text{scene}_1, x^\text{scene}_2, \ldots, x^\text{scene}_{T^\text{scene}}] \in \mathbb{R}^{T^\text{scene} \times 768},
\end{equation}
where $x^\text{scene}_t$ denotes the visual representation of the $t$-th sampled temporal position.

The extracted sequence of scene features is projected into a latent space and processed by a Mamba-based encoder~\cite{gu2023mamba}. In the main configuration, the temporal encoder has a hidden dimension of $128$ and consists of two Mamba layers with a state dimension of $64$, a one-dimensional convolution kernel of size $3$, and an expansion factor of $1$. Given the projected sequence $H^\text{scene}=[h^\text{scene}_1,h^\text{scene}_2,\ldots,h^\text{scene}_{T^\text{scene}}]$, each Mamba layer applies \gls{LN}, sequence mixing, dropout, and a residual connection:
\begin{equation}
    H{^\text{scene}}^{(l+1)} =
    H{^\text{scene}}^{(l)} + \text{Dropout}
    \left(
    \text{Mamba}\left(\text{LN}(H{^\text{scene}}^{(l)})\right)
    \right).
\end{equation}

The output sequence is aggregated using masked mean pooling to obtain a single scene representation, which is passed to the same \gls{MLP} classification head as that used in the face model. The output of the penultimate \gls{FCL} is used as the scene classification features, while the temporal hidden states is used as the temporal scene features.

\subsection{Multimodal Fusion Model}
Let $\mathcal{M}\subseteq\{\mathrm{audio},\mathrm{text},\mathrm{face},\mathrm{scene}\}$ denote the set of active modalities. For each modality $m$, we use a vector representation and, when available, the corresponding unimodal logits $\ell_m$. Temporal representations are converted into compact vectors using summary statistics. The best configuration uses:
\begin{equation}
    s_m=[\mu_m,\sigma_m,\mu_m^\Delta,\sigma_m^\Delta],
\end{equation}
where $\mu_m$ and $\sigma_m$ denote the mean and standard deviation of the temporal features, while $\mu_m^\Delta$ and $\sigma_m^\Delta$ denote the mean and standard deviation of their first-order temporal differences. The input to the fusion model is denoted by $u_m$. For example, when the unimodal logits are concatenated with the temporal statistics, $u_m=[s_m;\ell_m]$.

\begin{figure}[ht]
    \centering
    \includegraphics[width=\linewidth]{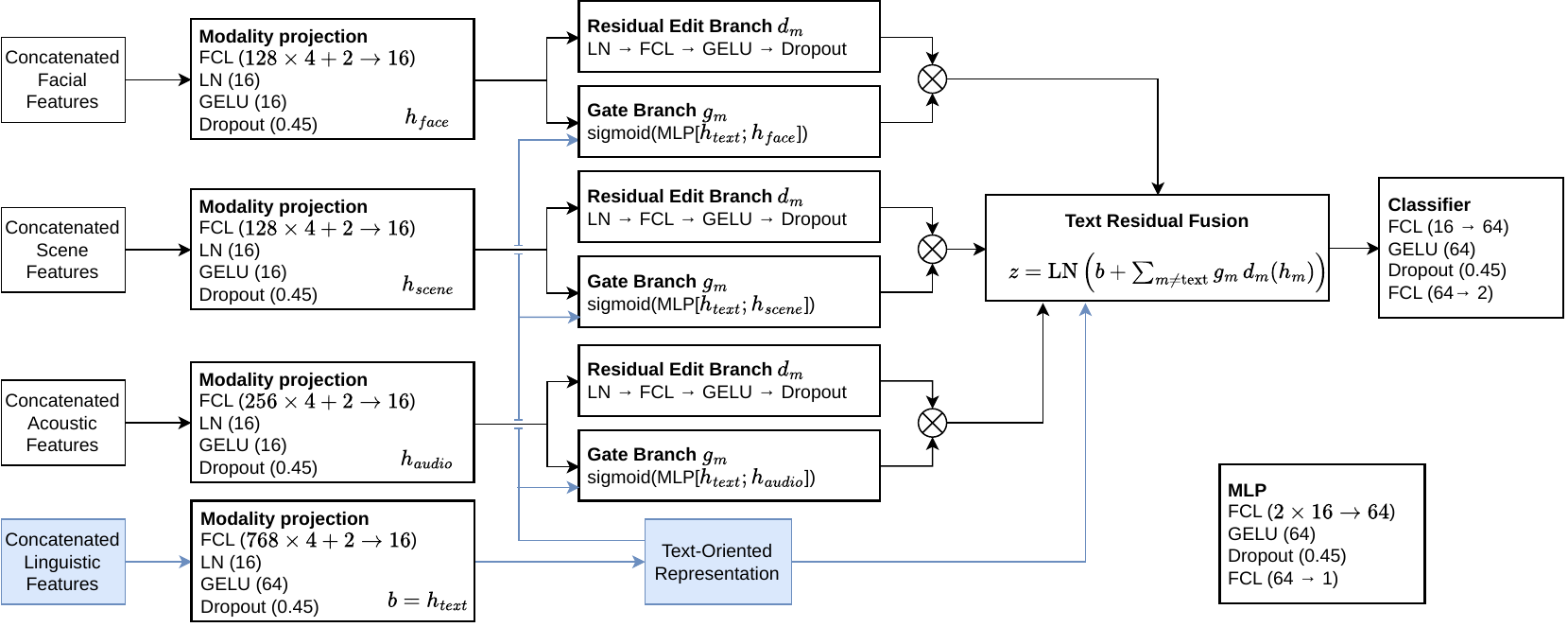}
    \caption{Pipeline of the Text Residual Fusion model.}
    \label{fig:pipeline}
\end{figure}

The \textbf{Text Residual Fusion} model uses text as the anchor modality, because it provides the strongest unimodal performance in our experiments. After modality-specific projections, the textual representation $b=h_{\mathrm{text}}$ is adjusted using gated residuals from the remaining modalities. For each non-text modality $m$, the gate is computed from the concatenated textual and modality-specific representations:
\begin{align}
    g_m = \sigma\left(\text{MLP} [b;h_m]\right),
    z = \text{LN}\left(b + \sum_{m\neq \text{text}} g_m\,d_m(h_m)\right),
\end{align}
where $d_m(\cdot)$ is a residual \gls{MLP}, $g_m$ is a learned gate. The gate controls the contribution of modality $m$ to the textual representation. The fused vector $z$ is then passed to a two-layer \gls{MLP} classifier. This design preserves the text as the primary source of information while allowing the audio, face, and scene modalities to provide input-dependent residual adjustments. The pipeline of the proposed model is shown in Figure \ref{fig:pipeline}.

\section{Experiments}

\subsection{Research Corpus}
\label{subsec:rs}

The \gls{BAH} corpus is the benchmark dataset for the \gls{AH} task of the 11th \gls{ABAW} Challenge. It was introduced for the multimodal \gls{AH} recognition in realistic digital behavior change scenarios~\cite{gonzalezbah}. Participants responded to a predefined set of questions intended to elicit neutral, positive, negative, willing, resistant, ambivalent, and hesitant responses during online interactions guided by virtual avatar~\cite{gonzalezbah}. 

The corpus contains $1427$ videos from $300$ participants, with a total duration of $10.60$ hours. It includes video-level and frame-level annotations, temporal boundaries of \gls{AH} episodes, aligned face regions, timestamped speech transcripts, and participant metadata~\cite{gonzalezbah}. Following the challenge protocol, \gls{AH} are treated jointly as a binary classification task. The dataset is divided at the participant level into the Train, Development, Public Test, and Private Test subsets. Performance is evaluated at the video-level using \gls{MF1}~\cite{gonzalezbah}.

\subsection{Experimental Results}

For all unimodal and multimodal experiments, we used Optuna~\cite{akiba2019optuna} to optimize the training procedure and model architecture. The final configurations (e.g., learning rate, dropout, hidden dimensions, batch size, and other parameters) were selected on the Development subset and evaluated on the Public Test subset.

As shown in Table~\ref{tab:bah_results}, text is the strongest unimodal modality, followed by audio. Both fusion models outperform all unimodal configurations on the Development and Public Test subsets. Text Residual Fusion achieves the highest Development MF1 of 76.13\%, the highest average MF1 across the Development and the Public Test subsets of 75.14\%, and the best Private Test MF1 of 78.24\%. On the Private Test subset, it outperforms the text model by 4.03\%. 

\begin{table}[tb]
\caption{Experimental results (MF1, \%) obtained by various configurations of the proposed approach. FM refers to flow matching. LS to label smoothing. TS to temporal smoothing.}
\label{tab:bah_results}
\centering
\resizebox{\textwidth}{!}{
\begin{tabular}{@{}llllllccc@{}}
\toprule
ID&Modality & Features & Temporal & Regular-  & Devel & Public Test & Average & Private Test \\
&& & model & ization& subset & subset &Devel/Public Test & subset \\
\midrule
1&Text & \multicolumn{2}{l}{RoBERTa-GoEmotions (freeze layers=4)} & -- &\textbf{72.55}& \textbf{72.10}&\textbf{72.33} &74.21\\
2& Audio & Wav2Vec2~\cite{baevski2020wav2vec2} & Bi-Mamba & TS &70.19 & 69.28 & 69.74 &-- \\
3&Face & EmoAffectNet~\cite{RYUMINA2022435}  & Transformer & FM  & 64.07 & 61.27 & 62.67 &-- \\
4&Scene& VideoMAE-v2 base~\cite{wang2023videomae} & MambaSSM & LS & 61.07 & 61.18 & 61.12&-- \\
\midrule
5&Text, Audio, Face, Scene&IDs 1, 2, 3 and 4 & Text Residual Fusion & LS & 76.13 &74.14 & 75.14 &78.24 \\

\bottomrule
\end{tabular}
}
\end{table}

\section{Conclusion}
This work presented a single text-centered multimodal approach that combines textual, acoustic, facial, and, scene features for video-level \gls{AH} recognition. The proposed Text Residual Fusion model uses text as the anchor modality and adds complementary information through gated residual adjustments.
The proposed fusion model achieves the best performance on the Private Test subset and the most consistent results across the evaluation subsets.


%
%
\bibliographystyle{splncs04}
\bibliography{main}
\clearpage
\end{document}

%% file: acronyms.tex
\newacronym{GRADA}{GRADA}{GRoup Affective Dynamics Analysis}
\newacronym{MLP}{MLP}{Multilayer Perceptron}
\newacronym{BAH}{BAH}{Behavioural Ambivalence/Hesitancy}
\newacronym{AH}{AH}{Ambivalence / Hesitancy}
\newacronym{MF1}{MF1}{Macro F1-score}
\newacronym{ABAW}{ABAW}{Affective \& Behavior Analysis in-the-Wild}
\newacronym{GELU}{GELU}{Gaussian Error Linear Unit}

\newacronym{M-LLM}{M-LLM}{Multimodal LargeLanguage Model}
\newacronym{LSTM}{LSTM}{Long Short-Term Memory}
\newacronym{ViT}{ViT}{Vision Transformer}
\newacronym{BERT}{BERT}{Bidirectional Encoder Representations from Transformers}

\newacronym{FCL}{FCL}{Fully-Connected Layer}
\newacronym{LN}{LN}{Layer Normalization}
\newacronym{SOTA}{SOTA}{State-of-the-Art}